\documentclass{article}
\usepackage{arxiv}
\usepackage[utf8]{inputenc} 
\usepackage[T1]{fontenc}    
\usepackage{hyperref}       
\usepackage{url}            
\usepackage{booktabs}       
\usepackage{amsfonts}       
\usepackage{nicefrac}       
\usepackage{microtype}      

\usepackage{natbib}
\usepackage{amsfonts}       
\usepackage{amsmath}
\usepackage{amssymb}
\usepackage{mathtools}
\usepackage{xcolor}
\usepackage{graphicx}
\usepackage{wrapfig}
\usepackage{caption}
\usepackage{booktabs}       
\usepackage{multirow}
\usepackage{dcolumn}
\usepackage[inline]{enumitem}
\usepackage{algorithm}
\usepackage[]{algpseudocode}
\usepackage{epsfig}
\usepackage{epstopdf}
\usepackage{times}
\usepackage{csquotes}
\usepackage{xr}
\usepackage{xr-hyper}
\usepackage{authblk}

\newcommand{\mc}[1]{\mathcal{#1}}

\DeclareMathOperator*{\argmax}{arg\,max}

\newcolumntype{.}{D{.}{.}{-1}}

\makeatletter
\def\blfootnote{\xdef\@thefnmark{}\@footnotetext}
\makeatother

\title{Relational Generalized Few-Shot Learning}

\author[1]{\textbf{Xiahan Shi}}
\author[1]{\textbf{Leonard Salewski}}
\author[1]{\textbf{Martin Schiegg}}
\author[2]{\textbf{Max Welling}}

\affil[1]{
	\textbf{Bosch Center for Artificial Intelligence}\protect\\
	Robert-Bosch-Campus 1\protect\\
	71272 Renningen, Germany\protect\\
	\url{www.bosch-ai.com}\protect\\
	$\:$}

\affil[2]{
	\textbf{University of Amsterdam}\protect\\
	Science Park 904\protect\\
	1098 XH Amsterdam}

\begin{document}

\maketitle
\rhead{Shi et al.: Relational Generalized Few-Shot Learning}

\begin{abstract}
Transferring learned models to novel tasks is a challenging problem, 
particularly if only very few labeled examples are available. 
Most proposed methods for this few-shot learning setup focus on discriminating 
novel classes only. Instead, we
consider the extended setup of \textit{generalized few-shot learning} (GFSL), 
where
the model is required to perform classification on the \textit{joint} label 
space
consisting of both previously seen and novel classes. 
We propose a graph-based
framework 
that
explicitly models relationships between \emph{all} seen and novel classes in 
the joint label space.
Our model \textit{Graph-convolutional Global Prototypical Networks} (GcGPN)  
incorporates these inter-class relations using graph-convolution in order to 
embed novel class representations into the existing space of previously seen 
classes in a globally consistent manner.
Our approach ensures both fast adaptation and global discrimination, which is 
the major 
challenge in GFSL.
We demonstrate the benefits of our model on two challenging benchmark datasets.
\end{abstract}


\section{Introduction}%
\label{sect:introduction}
\blfootnote{Published at the 31\textsuperscript{st} British Machine Vision Conference 2020.}
\blfootnote{The paper and a video recording of the talk can be found \href{https://bmvc2020-conference.com/conference/papers/paper_0220.html}{here}.}
\blfootnote{Correspondence to \texttt{xiahan.shi@de.bosch.com}}

\begin{figure}[h]
	\centering
	\includegraphics[width=0.99\textwidth]{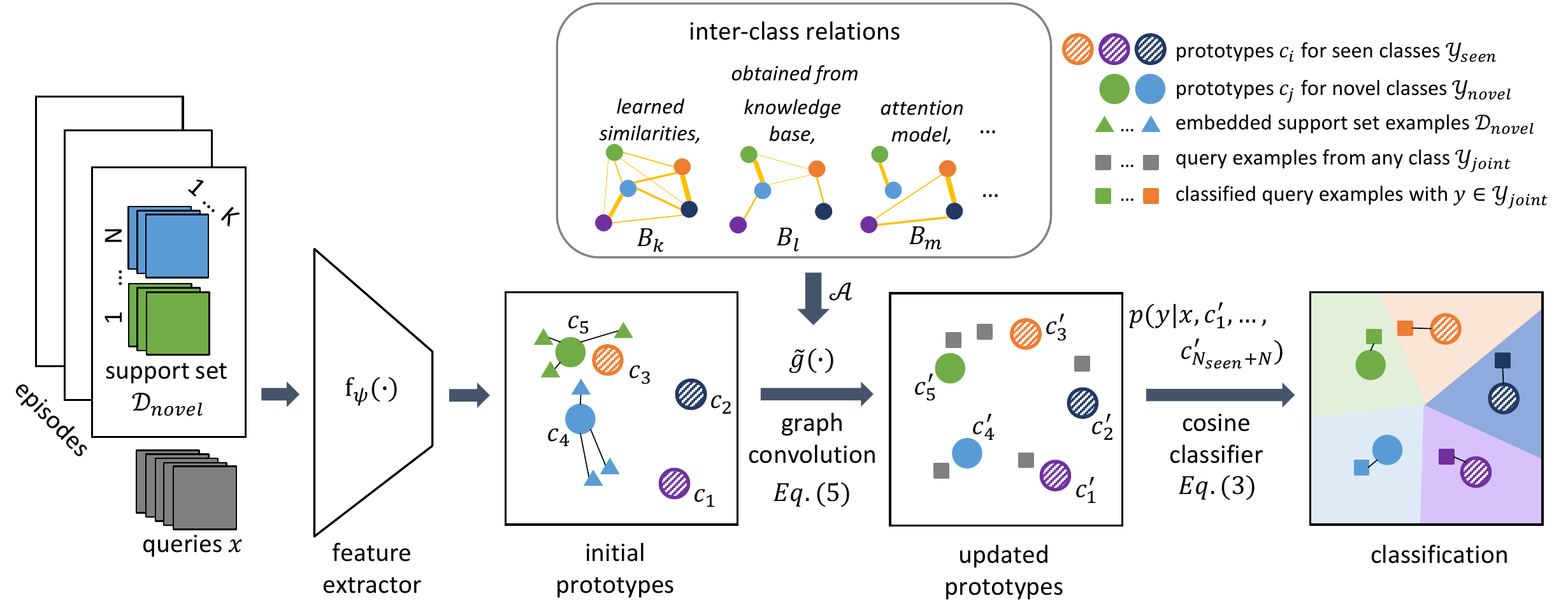}
	\vspace{0.5em}
	\caption{\small
		\textbf{Our framework:}
		In an $N^+$-way $K$-shot episode, the task is to discriminate the joint 
		label space $\mc Y_{\text{joint}} = \mc Y_{\text{seen}} \cup \mc 
		Y_{\text{novel}}$. While $\mc Y_{\text{seen}}$ contains (a large number 
		of) seen classes, $\mc Y_{\text{novel}}$ consists of $N$ previously 
		\emph{unseen} classes with only $K$ labeled \emph{support set} examples 
		per class.
		The goal of GFSL is to perform well across a series of such $N^+$-way 
		$K$-shot episodes with \emph{varying} $\mc Y_{\text{novel}}$. 
		GcGPN addresses this challenge by explicitly modeling relationships 
		between \emph{all} classes in $\mc Y_{\text{joint}}$ as graphs, where 
		edges (yellow) represent inter-class relations:
		First, a feature extractor $f_\psi$ maps all support set and query examples to a 
		feature space, 
		where classes are represented by prototypes. While novel class 
		prototypes (solid circles) are initialized by averaging the 
		corresponding support set feature representations (triangles), initial 
		seen class prototypes (shaded circles) are modeled as learnable 
		parameters.
		Then, GcGPN employs graph-convolutions to propagate information among 
		classes 
		according to the given inter-class relations, resulting in jointly 
		updated prototypes. 
		At last, the queries (grey rectangles) are classified according to 
		their cosine similarity to all prototypes. 
		The model is trained end-to-end in a meta-learning setup.}%
	\label{fig:gcgpn}
	\vspace*{-2.0ex}
\end{figure}

Few-shot learning (FSL)~\cite{thrun1996learning,miller2000learning,bart2005cross,fei2006one} 
is inspired by the human ability to learn new concepts 
from very few or even only one example. This extreme low-data setup is particularly 
challenging for deep neural networks, which require large amounts of data to ensure
generalization.
FSL has mostly been approached from the meta-learning perspective, 
focusing on the problem setup of \emph{$N$-way $K$-shot classification}.
For each $N$-way $K$-shot task, the model has to discriminate $N$ novel few-shot 
classes with only $K$ labeled examples available per class. 
Unlike in standard transfer learning,
meta-learning requires the model to adapt well across a \emph{series of various} 
previously unknown tasks instead of a fixed, \emph{specific} target task.
Therefore, efficient \emph{on-the-fly} model adaptation based on very few examples 
is at the core of most FSL 
models~\cite{vinyals2016matching,snell2017prototypical,finn2017model,ravi2016optimization,garcia2017few,gidaris2018dynamic,li2019finding}.

However, this FSL problem setup focuses only on discriminating novel classes from each other 
and offers no incentive for the model to remember classes previously seen during training or to 
maintain a globally consistent label space. 
However,
we would like the model to \emph{incorporate} few-shot novel classes 
into the label space of previously seen classes. 
This leads us to an extended scenario called \emph{generalized few-shot learning} (GFSL), 
where the model has to discriminate the \emph{joint} label space consisting of both 
previously seen and novel classes. 
This terminology is adopted from zero-shot learning (ZSL) and generalized 
zero-shot learning (GZSL), where novel classes come with no labeled examples at all and classification relies on side information such as attributes or semantic label embeddings~\cite{akata2016label, xian2017zero}.
It is a well-known observation that many ZSL models fail dramatically at 
discriminating the joint label space (GZSL) despite good performance 
on novel label space (ZSL)~\cite{chao2016empirical,xian2017zero}. 
Similarly, GFSL is a more challenging task compared to FSL due to 
the trade-off between fast adaptation to novel classes and maintaining 
a global consistency across the joint label space.

We address the GFSL problem setup by explicitly modeling inter-class 
relationships as a weighted graph.
We propose the \emph{Graph-convolutional Global Prototypical Network} (GcGPN)
which models representative 
prototypes for all novel and seen classes \emph{jointly}. 
In particular, the prototypes are updated by 
graph convolutional operations~\cite{kipf2016semi} according to the relationship graph. 
Fig.~\ref{fig:gcgpn} provides an illustration of our approach.
To summarize, our \textbf{main contributions} are: We propose the first 
flexible framework for relational GFSL that 
\begin{enumerate}[label=(\arabic*)]
	\item considers an \emph{arbitrary weighted graph} describing relations between classes (from any source of side information, attention mechanism or other similarity measures),
	\item applies graph-convolution for modeling prototypes and allows 
	for end-to-end~learning,
	\item accommodates previous (G)FSL methods~\cite{snell2017prototypical,gidaris2018dynamic} as special cases and
	\item achieves state-of-the-art performance on GFSL tasks.
\end{enumerate}

\section{Related Work}



\textbf{Few-shot learning (FSL)} has been approached from different perspectives including 
mimicking the human learning behavior by modeling high-level concepts~\cite{lake2015human}, 
learning similarity measures~\cite{koch2015siamese}
and extending deep neural networks 
by an external memory module to allow for direct incorporation of few-shot examples~\cite{santoro2016meta, kaiser2017learning}. 
Moreover, recent \emph{meta-learning} approaches focus on the $N$-way $K$-shot setup and can 
be divided in two categories: Optimization-based methods~\cite{finn2017model,ravi2016optimization,antoniou2018train}
rely on a meta model that learns an optimal strategy which is carried out by an inner model to  
efficiently adapt to varying novel tasks. Distance-based methods such as \textit{Matching Networks}~\cite{vinyals2016matching} 
and \textit{Prototypical Networks}~\cite{snell2017prototypical} perform nearest-neighbor-based 
classification with a learned distance measure.
Despite its simplicity, the method in~\cite{snell2017prototypical} achieves 
excellent
performance and has inspired extensions which parameterize the 
distance measure or the prototype mechanism in a more flexible 
way~\cite{sung2018learning}.

\textbf{Generalized few-shot learning (GFSL)} in the context of meta-learning is not yet a  
well-studied setup. Alternatives to meta-learning include matching seen and novel feature spaces~\cite{verma2017simple,schonfeld2018generalized}, modeling the global class structure 
in the joint label space~\cite{lifew}, obtaining transferable features from a
hierarchy between seen and novel classes via clustering~\cite{li2019large}, 
propagating labels from class-level to instance-level 
graphs~\cite{zhang2019tgg},
 and learning generative models to synthesize additional
image features for the few-shot classes~\cite{xian2019f} or additional 
samples~\cite{li2019few}. These methods, however, require knowledge about all 
novel 
classes \emph{a priori}, e.g., by leveraging a pre-trained feature space or graph 
construction, which 
is in contrast to the meta-learning setup where
the model must be able to adapt on-the-fly to unknown and varying novel classes.
The most relevant work to our target setup is \textit{Dynamic FSL 
without Forgetting}~\cite{gidaris2018dynamic} (DFSLwoF),
which utilizes within a meta-learning setup an attention-based weight generator 
for novel classes to extend the classifier 
from seen classes to the joint label space. Its connections to our work is 
discussed in detail in Section~\ref{PN+}. Their work is extended 
in~\cite{gidaris2019generating} by a graph neural network (GNN) based
denoising autoencoder to regularize the class prototypes,
where the underlying graph is a special case of our fully connected setup.
Further, the GNN layer consists of a separate neighbor-aggregation block
based on Relation Networks~\cite{santoro2017simple} and an update-block
to combine the prototypes with their respective neighbor-information.
In contrast, our method uses a simpler model structure while
providing a more general framework to include side information,
and is trained end-to-end instead of the two-stage training procedure
in~\cite{gidaris2019generating,santoro2017simple}.
An orthogonal approach has been proposed in~\cite{qiao2018few},
where a pre-trained network can be extended to additional few-shot classes
by predicting the final-layer parameters from the activations.
This method is most notably useful when working with an existing,
very strong model where further training is hard to realize.

\textbf{Side Information} plays a crucial role in zero-shot learning (ZSL), where no labeled examples are available for novel classes at all. In particular, ZSL methods typically build on side information from
knowledge graphs~\cite{miller1995wordnet}, semantic word embeddings~\cite{mikolov2013distributed,pennington2014glove} or visual attributes~\cite{duan2012discovering}. 
Generalization can be achieved by relating visual image features to semantic side information either through
a learned mapping or a joint embedding space~\cite{frome2013devise,norouzi2013zero,akata2016label,schonfeld2018generalized}.
Furthermore, graph-convolutions can be applied to distill information from relational knowledge graphs 
and class embeddings to predict the classifier weights for unseen classes~\cite{wang2018zero}.
Apart from ZSL, a range of FSL methods exist that incorporate side information,
e.g., to regularize the feature space with textual embeddings for alignment on 
a distributional level~\cite{tsai2017learning}, to use
attention mechanisms for synthesizing additional training 
examples for few-shot classes~\cite{tsai2017improving}, or to match a visual 
classifier with a knowledge-based representation~\cite{peng2019few}.
Besides the low-shot data regime, relational information has 
also been used to improve loss functions for deep learning in general~\cite{xu2017semantic}.

\textbf{Graph-convolutional networks (GCN)}~\cite{duvenaud2015convolutional,Defferrard2016ConvolutionalNN,kipf2016semi} are a powerful tool to jointly learn node representations for inherently graph-structured data 
such as items in recommender systems or users of social 
networks~\cite{berg2017graph}. Graph-based methods
have been applied to FSL~\cite{garcia2017few,kim2019edge} by representing \emph{image 
instances} with graph nodes   
in an $N$-way $K$-shot classification setup. In contrast, we represent \textit{classes} by graph nodes with a GFSL setup.
Class-level graph-convolution has been used in a similar way for ZSL~\cite{wang2018zero}.
Alternatively, GCNs may exploit the manifold structure in the data to propagate labels from labeled to unlabeled images by using edge weights that depend on learned distances in the feature space~\cite{liu2018learning}.


%
%

\section{(Generalized) Few-Shot learning}



\paragraph{Few-shot learning (FSL)}
We consider $N$-way $K$-shot classification, which is the most widely 
studied problem setup for FSL. The classifier has to perform a series 
of $N$-way $K$-shot tasks, where each task consists of $N$ previously 
unseen, novel
classes with $K$ labeled examples each (usually $K\leq5$). More precisely, 
let $\mc Y_{\text{novel}}$ denote the novel class label space with $|\mc Y_{\text{novel}}|=N$, 
and let $\mc D_{\text{novel}} = \cup_{n=1}^{N} {\left\{(x_{n,k},y_n)\right\}}_{k=1}^{K}$
denote the \emph{support set}, where $x_{n,k}$ is the $k$-th 
labeled example of the class with label $y_n$. For a new query example $x$, the FSL prediction
is given by
\begin{equation}
\hat{y} = \argmax_{y \in \mc Y_{\text{novel}}} p_\psi( y | x, \mc D_{\text{novel}}).
\label{eq:fsl}
\end{equation}
$N$-way $K$-shot classification considers FSL from a \emph{meta-learning} 
perspective. Unlike in standard transfer learning, the goal is not 
to generalize to a \emph{specific} novel label space but to adapt and
perform well across a series of \emph{various} novel label spaces presented at test time. 
Therefore, most FSL methods adopt \emph{episodic training}~\cite{vinyals2016matching}, 
where a new $N$-way $K$-shot task gets randomly sampled from a larger training set 
in every episode. 
This involves randomly selecting $N$ simulated\footnote{
	The ``novel'' classes at training time are randomly sampled from the 
	\emph{training} classes in order to simulate the test time setup, but they 
	are \emph{disjoint to the real novel classes at test time}.}
novel classes and sampling
$K$ support set examples per class along with a batch $\mathcal{B}$ of query examples.
The loss on this batch is given by
$\frac{1}{|\mathcal{B}|} \sum_{(x,y)\in \mathcal{B}} - \log p_\psi(y | x, \mc D_{\text{novel}})$,
which is used to train the model parameters $\psi$.

An FSL model is only concerned with discriminating the novel label space since 
all test time queries, by design of the task setup, come from one of the novel classes.
Hence the $\argmax$ in eq.~\eqref{eq:fsl} is only over $\mc Y_{\text{novel}}$. Classes seen 
during training no longer play a role at test time. 
This setup emphasizes fast adaptation to varying new tasks but does not 
encourage the model to accumulate knowledge, which may not always be very practical. Many real-world applications require the model to incorporate novel few-shot classes into the existing space of seen 
classes while maintaining global discrimination.
Therefore, we consider the extended setup of \emph{generalized few-shot learning} (GFSL) with test time queries that may come 
from both novel and seen classes.

\paragraph{Generalized few-shot learning (GFSL)}
In generalized $N^{+}$-way $K$-shot classification, the model has to discriminate the 
joint label space $\mc Y_{\text{joint}} = \mc Y_{\text{seen}} \cup \mc Y_{\text{novel}}$ consisting of 
the novel few-shot classes and all previously seen training classes.
We denote the training set by
$\mc D_{\text{seen}} = \cup_{n=1}^{N_{\text{seen}}} {\left\{ (x_{n,k},y_n) \right\}}_{i=1}^{K_n}$,
where $N_{\text{seen}}$ is the number of training classes and $K_n$ is the number of labeled 
examples available for class $y_n \in \mathcal Y_{\text{seen}}$. In general, $N_{\text{seen}}\gg N$ and $K_n \gg K$. 
For a new query $x$, a GFSL model performs 
\begin{equation}
\hat{y} = \argmax_{y \in \mc Y_{\text{joint}}} p_\psi( y | x, \mc D_{\text{novel}}).
\label{eq:gfsl}
\end{equation}
In contrast to eq.~\eqref{eq:fsl}, the $\argmax$ is over $\mc Y_{\text{joint}}$ instead of  $\mc Y_{\text{novel}}$ 
since a query $x$ may come from any of the seen and novel classes.
In particular, GFSL requires discrimination of a much larger label space than FSL 
($N^{+} := N + N_{\text{seen}}$ instead of~$N$). 
In addition, the model has to maintain a globally consistent joint label space 
while, at the same time, achieve fast adaptation 
to novel classes based on very few examples.
In general, we cannot expect FSL models to perform well in GFSL since there is 
no explicit reward to remember the training classes and learn a 
well-separated joint label space.



\section{Graph-convolutional Global Prototypical Networks}

We propose \emph{Graph-convolutional Global Prototypical Networks} (GcGPN) to perform relational GFSL.
The key idea is to explicitly model and incorporate the relationships among \emph{all}
(e.g., including seen and novel) classes 
through a weighted graph when learning class representations (so-called \emph{prototypes}).
This addresses the challenge of GFSL to maintain \emph{global} consistency and discrimination 
when incorporating novel classes into an existing space of seen classes.

\subsection{GcGPN: Model Overview}%
\label{sec:GcGPN_model_overview}

Fig.~\ref{fig:gcgpn} illustrates the full pipeline of our method. 
First, GcGPN maps all support set and query examples into a $d$-dimensional feature space
by a feature extractor $f_\psi(\cdot)$. Next, initial prototypes 
$c_n \in \mathbb{R}^d, \: n=1,\ldots,N_{\text{seen}}+N$, are computed for all classes. 
While seen class prototypes are learned as model parameters, novel class prototypes 
are initialized on-the-fly since the novel label space varies at test time.
The novel initial prototypes are given by the per-class average 
$c_n = \frac{1}{K} \sum_{i=1}^{K}\bar{z}_{n,i}$ of the normalized support set examples 
$\bar{z}_{n,k} = \frac{f_\psi(x_{n,k})}{||f_\psi(x_{n,k})||}$,
$n=N_{\text{seen}}+1,\ldots,N_{\text{seen}}+N$, as in~\cite{snell2017prototypical, gidaris2018dynamic}.
Then, a graph-convolution block $\tilde g(\cdot)$ 
as defined below 
updates these 
node initializations jointly according to the inter-class 
relationships specified in the edge weights. The updated prototypes 
$c'_n, \: n=1,\ldots,N_{\text{seen}}+N$, are then adapted representations of the joint 
label space of the $N^{+}$-way $K$-shot task at hand. Finally, 
the predicted class probabilities for a query $x$ are obtained from its 
cosine similarities between its feature representation and the updated prototypes:
\begin{equation}
p(y=n|x, c'_1,\ldots,c'_{N_{\text{seen}}+N}) =
 \frac{\exp\left(\tau \cos(f_\psi(x), 
 c'_n)\right)}{\sum_{m=1}^{N_{\text{seen}}+N} \exp\left(\tau \cos(f_\psi(x), 
 c'_m)\right)},
\label{eq:GcGPN}
\end{equation}
where $\tau$ is a learnable temperature. In~\cite{gidaris2018dynamic}, 
this is referred to as the \textquote{cosine classifier}. We adopt it since it 
was found to be preferable over the originally proposed L2 distance~\cite{snell2017prototypical}
when combining existing and novel classes~\cite{bauer2017discriminative,gidaris2018dynamic}.
To train the model, we use the cross-entropy loss on eq.~\eqref{eq:GcGPN}. 
Note that the sum in eq.~\eqref{eq:GcGPN} is over \emph{all} class prototypes 
in the \textit{joint} label space, which is in accordance with eq.~\eqref{eq:gfsl} 
and differs from the FSL objective. Further, we apply episodic training for GFSL:
For each episode, $N$ out of $N_{\text{seen}}$
training classes are sampled to act as novel classes and the remaining 
$N_{\text{seen}}-N$ are treated as the label space of seen classes.
Contrary to an FSL episode, the GFSL query set $\mathcal{Q}$ must also contain examples 
from the seen classes, thus rewarding the model for maintaining global discrimination 
instead of focusing only on the novel classes. In every episode, 
the gradient of the loss is computed and all learnable parts of the model 
get updated including the parameters $\psi$ of the feature extractor, the initial prototypes $c_n,\:n=1,\ldots,N_{\text{seen}}$ of seen classes, trainable components of the graph-convolution block $\tilde g(\cdot)$ and the classifier temperature $\tau$. 
Unlike previous work~\cite{gidaris2018dynamic}, our model does not require 
multi-stage training but trains all parts of GcGPN jointly.

\paragraph{The graph-convolution block}
The graph-convolution block $\tilde g(\cdot)$
is at the core 
of GcGPN. To recap, a graph-convolution block~\cite{kipf2016semi} consists of $L$ graph-convolution layers $g(\cdot)=g^L(g^{L-1}(\ldots(g^1(\cdot))))$
on a graph of $V$ nodes, which is given in its general form by 
\begin{equation}
X^{(l+1)} = g^l(X^{(l)}) = \rho(\sum_{B\in \mc A} B X^{(l)} \theta_{B}^{(l)}), 
\:
\label{eq:gconv}
\end{equation}
where $l\in \{1,\ldots,L\}$ indexes the layer of the block, $X^{(l)}$ is the $(V 
\times 
d_l)$-dimensional input matrix containing 
$d_l$-dimensional node features in its rows,
$\mathcal A$ denotes a set of $(V \times V)$-dimensional linear node operators such as the adjacency or weight matrix of the graph, 
$\theta_{B}^{(l)}$ with $B \in \mathcal A$ denotes a $(d_l \times d_{l+1})$-dimensional matrix containing 
learnable parameters of the $l$-th layer and $\rho(\cdot)$ is a non-linearity. 
For example, if $B$ is the adjacency matrix of the graph, the local convolutional operation 
$BX^{(l)}$ computes for each node the sum of its neighbors.

In our GFSL model, eq.~\eqref{eq:gconv} is applied to the class prototypes to model relations between them. More precisely,
let $C$ denote the $((N_{\text{seen}}+N) \times d)$-dimensional matrix, where the $n$-th row contains
the initial prototype $c_n$ of the $n$-th class. Further, let the operator entries $B_{m,n}$ encode 
a similarity score between classes represented by $c_m$ and $c_n$. 
Then, $L$-layered graph-convolution block takes $X^{(0)}:=C$ as input and computes the 
updated class prototypes as $C'= g(C) = g^L(\ldots g^1(C))$ according to 
eq.~\eqref{eq:gconv}.

Note that graph-convolution can be interpreted as performing two steps to update 
a class prototype: First, a weighted average of similar prototypes is computed with weights given in the 
\emph{convolution operator} $B$ and second, 
a non-linear \emph{post-convolution transform} is applied given by $\theta_B^{(l)}$ and $\rho(\cdot)$.
The first part models interactions among classes and operates only on node-level, 
while the second part operates only on feature-level by applying the same transform to all classes.

The general graph-convolution definition from eq.~\eqref{eq:gconv} operates in Euclidean spaces. We adopt the 
graph-convolution block to be consistent with cosine similarities as used in eq.~\eqref{eq:GcGPN} by intermediate 
normalizations $\bar x = \frac{x}{\|x\|}$ to keep the prototypes at unit length. Our graph-convolution block is thus defined by
$C'=\tilde{g}(C)=\tilde g^L(\ldots\tilde g^1(C))$ with
$\tilde{g}^l(C^{(l)}) = \rho( \sum_{B\in \mc A} s_{B} \overline{ B 
\bar{C}^{(l)} \theta^{(l)}_{B} }),\ l=1,\ldots,L$,
where scalars $s_B$ are introduced to trade-off between different operators in 
$\mathcal A$. 
The relational information between the classes is modeled by the operators 
$B\in \mathcal A$ and we call these the \emph{semantic} operators.


\paragraph{Graph-convolutional operators:} 
In typical applications for graph convolutional networks such as recommender systems or 
social network analysis, the adjacency matrix is a popular choice~\cite{kipf2016semi}.
Inter-class relationship modeling suggests to more generally use a weighted graph,
where entries express some notion of similarity.
In general, there are several possible strategies to define the operator 
entries $B_{n,m}$: 

\begin{enumerate}[label=(\arabic*)]
\itemsep-0.2em
\item Any distance/similarity on the prototype space such as 
L2 distance or cosine similarity.
\item Learned similarities, either using a standard measure in a learned transformed 
space or learning a flexible transform of the element-wise absolute differences as done in~\cite{garcia2017few}.
\item Similarities within a learned key space as proposed in~\cite{gidaris2018dynamic}. This means learning a key $k_n$ for each class $n$ and obtaining inter-class similarities by matching the corresponding keys in the key space.
\item Side information 
from external sources of information such as knowledge graphs or semantic 
models. 
	This can be extracted as relational scores (e.g. shortest-path distance between two class names in an ontology~\cite{wang2018zero}) directly or obtained from per-class embeddings such as word vectors or attributes~\cite{akata2016label} by computing pairwise similarity.
\end{enumerate}

Note that 
also sparse graphs (as arising from e.g. sparse knowledge graph structure, 
operator thresholding, adjacency) are covered and 
higher-order structures are easily incorporated, e.g. by 
adding higher-order versions of the adjacency matrices to $\mathcal{A}$ or using
similarity scores that already contain higher-order information such as the 
path similarity in WordNet~\cite{wang2018zero}.

Due to the multi-operator design, our model can naturally combine multiple of the 
above strategies, resulting in a general and flexible framework for relational GFSL.

\paragraph{Post-convolutional transforms:} The parameters $\theta_B$ are another crucial component of the model. 
Since the output of $\tilde g(\cdot)$ are updated prototypes, we choose $\theta_B$ to preserve the
dimensionality. Using a learnable quadratic weight matrix is the most straight-forward
approach, although restricting $\theta_B$ similar to~\cite{gidaris2018dynamic} is also 
a competitive option.

\subsection{Generalization of existing approaches}%
\label{PN+}

There is a naive extension of the state-of-the-art FSL method 
\textit{Prototypical Networks}~\cite{snell2017prototypical} (PN) to the GFSL 
setup: For a readily trained PN, 
seen class prototypes can be obtained as feature averages over all available 
training examples 
for that class\footnote{%
	This is in contrast to our method, where prototypes are learnable 
parameters, \textit{initialized} with an 
	average over the support points.
}, which can then be used to perform GFSL tasks.
This extension referred to as PN$^+$ corresponds to the assumption that
the learned feature extractor and prototype mechanism would naturally
generalize over the joint label space.
Thus, there is no explicit inter-class dependency modeling, 
which is equivalent to setting all $B\in\mathcal A$, $\rho(\cdot)$, $s_B$, $\theta_B^{(l)}$ to identity matrices or functions in our GcGPN framework.
We will observe in the experiments later that this assumption is not appropriate.



The model in~\cite{gidaris2018dynamic} addresses GFSL successfully by an 
attention-based weight generator 
that computes classifier weights $w^*$ for novel classes based on their support sets and 
their similarities to seen classes. 
Both our model and theirs utilize a cosine classifier. However, while the cosine classifier 
in our model operates on representative prototypes in the feature space, 
theirs operates in the weight space and computes cosine similarities between 
seen class weights and transformed support set image features.
The \textit{Average Weight Generator} variant in~\cite{gidaris2018dynamic} can be recovered in our framework 
by using a GcGPN with prototype 
initialization as in sec.~\ref{sec:GcGPN_model_overview} and one graph-convolution layer ($L=1$) with 
$\mathcal A=\{\hat B_1,\hat B_2\}$ containing two block-structured operators
\begin{equation}
	\begin{gathered}
	\hat B_1 =\left(\begin{array}{ll}
	I_{N_{\text{seen}} \times N_{\text{seen}}} & 0_{N_{\text{seen}} \times N} \\
	0_{N \times N_{\text{seen}}} & 0_{N \times N}
	\end{array}\right), \qquad
	\hat B_2 = \left(\begin{array}{ll}
	0_{N_{\text{seen}} \times N_{\text{seen}}} & 0_{N_{\text{seen}} \times N} \\
	0_{N \times N_{\text{seen}}} & I_{N \times N}
	\end{array}\right)
	\end{gathered}%
	\label{eq:auxiliary_ops}
\end{equation}
with identity matrix on the seen- and novel-class blocks respectively and zeros 
elsewhere. This corresponds to not modeling inter-class relations at all. 
$\theta_{\hat B_1}$ is the identity matrix and $\theta_{\hat B_2}$ is a learnable diagonal matrix.
The \textit{Attention Weight Generator} variant in~\cite{gidaris2018dynamic}
can be recovered by adding one more operator to $\mathcal A$, 
whose lower-left block contains attention weights that are 
obtained by matching the respective classes in a learned key space 
(see semantic operator option 3 above).
This corresponds to an underlying relational graph with weighted directed edges from seen to novel classes,
such that novel prototypes do not only depend on the support set but also on similar seen classes.

To summarize, GcGPN generalizes over~\cite{gidaris2018dynamic} in several respects: 
\begin{enumerate*}[label=\textit{(\roman*)}]
	\item We use a fully connected graph, allowing not only relations from seen to novel classes but among \emph{all} classes (i.e., operators in $\mathcal A$ may be full matrices); 
	\item our framework accommodates any kind of similarity modeling (not only attention matching) 
	and offers a natural way to combine multiple strategies (see semantic 
	operators 
	(1)--(4));
	\item more general post-convolution transforms and layer stacking ($L\geq 1$) result in a more flexible joint model for class prototypes;
	\item all parameters can be trained end-to-end through a GFSL objective, thus does not require the 2-stage training procedure from~\cite{gidaris2018dynamic}.
\end{enumerate*}

\section{Experiments}%
\label{sec:experiments}

We evaluate 
our method
on two widely used benchmark datasets. 
First, we use the FSL benchmark dataset 
\textbf{miniImageNet}~\cite{vinyals2016matching}, 
which consists of 100 classes and 600 images per class. 
We adopt the split specified in~\cite{ravi2016optimization} to obtain 64 seen,
16 novel validation and 20 novel test classes. 
To obtain training, validation and test sets for the \textit{seen} class label space,
we further follow the approach in~\cite{gidaris2018dynamic}.
We enrich this dataset with side information based on conceptual semantics and lexical relations
by mapping class names into the
ontology \textit{WordNet}~\cite{miller1995wordnet}. 
In particular,
we use \textit{WordNet} path similarities~\cite{pedersen2004wordnet} 
between class labels, which are scores based on the shortest 
path distances between words 
in the taxonomy.
Second, we evaluate our method on \textbf{Caltech-UCSD Birds-200-2011}~(\textit{CUB})~\cite{WahCUB_200_2011}, 
which is widely used for ZSL. 
This dataset contains 11,788 images across 200 classes of different bird species. 
Each class has 312 annotated continuous attributes describing visual characteristics of the 
respective bird species.
We follow the standard split used in ZSL~\cite{MorgadoCVPR17} to obtain 150 seen and 50 novel test classes. 
Further, we randomly select 20 from the 150 seen classes for validation. 
For each seen class, 25\% of the images are hold out as seen class test set 
and 10\% as seen class validation set.
In this dataset, we obtain edge weights by computing pairwise cosine similarities between class attributes.
These 
semantic operators $\tilde B$, where class similarities are used as edge 
weights, are depicted in the supplementary.

%

\begin{table}[t]
\centering
\setlength{\tabcolsep}{3pt}
\resizebox{0.78\linewidth}{!}{
	\begin{tabular}{ l c c c c c c }
		\toprule[0.25ex]
		\multirow{2}{*}{}                                            
		&           & 
		FSL            &                                   
		\multicolumn{4}{c}{GFSL}                                   \\
		\cmidrule(r){3-3}\cmidrule(r){4-7}\multicolumn{1}{l}{1-shot} & 
		Seen-Seen & 
		Novel-Novel    & Joint-Joint             & Seen-Joint & 
		Novel-Joint               & H-Mean                    \\
		\midrule
		PN$^+$~\cite{snell2017prototypical}                          & 
		54.02$\pm$0.46     & 53.88$\pm$0.78 & 27.02$\pm$0.23          & 
		54.02$\pm$0.46      & \hphantom{0}0.02$\pm$0.01 & 
		\hphantom{0}0.04$\pm$0.03 \\
		DFSLwoF~\cite{gidaris2018dynamic}                            & 
		69.93$\pm$0.41     & 55.80$\pm$0.78 & 49.42$\pm$0.41          & 
		58.54$\pm$0.43      & 40.30$\pm$0.74            & 
		46.95$\pm$0.55            \\
		GcGPN                                                        & 
		63.68$\pm$0.42     & 55.67$\pm$0.73 & 46.82$\pm$0.41          & 
		51.08$\pm$0.46      & 42.57$\pm$0.72            & 
		45.68$\pm$0.48            \\
		GcGPN-aux                                                    & 
		68.39$\pm$0.40     & 56.59$\pm$0.75 & 49.66$\pm$0.39          & 
		58.16$\pm$0.44      & 41.16$\pm$0.69            & 
		47.51$\pm$0.51            \\
		GcGPN-split                                                  & 
		68.26$\pm$0.42     & 55.68$\pm$0.76 & 49.60$\pm$0.41          & 
		55.22$\pm$0.47      & 43.98$\pm$0.76            & 
		48.13$\pm$0.49            \\
		GcGPN-aux-split                                              & 
		68.13$\pm$0.43     & 60.40$\pm$0.71 & \textbf{51.63$\pm$0.41} & 
		54.68$\pm$0.46      & 48.59$\pm$0.72            & 
		\textbf{50.83$\pm$0.45  
		} \\
		
		GcGPN-cos-aux & 69.86$\pm$0.41 & 54.00$\pm$0.77 & 47.94$\pm$0.40 & 62.39$\pm$0.45 & 33.50$\pm$0.67 & 42.88$\pm$0.59\\
	
		\midrule[0.25ex]
		\multicolumn{1}{l}{5-shot}                                   & 
		Seen-Seen & 
		Novel-Novel    & Joint-Joint             & Seen-Joint & 
		Novel-Joint               & H-Mean                    \\
		\midrule
		PN$^+$~\cite{snell2017prototypical}                          & 
		60.42$\pm$0.45     & 70.84$\pm$0.66 & 31.70$\pm$0.25          & 
		60.41$\pm$0.45      & \hphantom{0}2.99$\pm$0.20 & 
		\hphantom{0}5.54$\pm$0.34 \\
		DFSLwoF~\cite{gidaris2018dynamic}                            & 
		70.24$\pm$0.43     & 72.59$\pm$0.62 & \textbf{59.08$\pm$0.40} & 
		59.89$\pm$0.47      & 58.26$\pm$0.68            & 
		\textbf{58.58$\pm$0.41  
		} \\
		GcGPN                                                        & 
		66.51$\pm$0.43     & 71.53$\pm$0.63 & 57.16$\pm$0.40          & 
		56.73$\pm$0.45      & 57.59$\pm$0.67            & 
		56.69$\pm$0.41            \\
		GcGPN-aux                                                    & 
		68.89$\pm$0.43     & 71.81$\pm$0.64 & 58.03$\pm$0.39          & 
		60.56$\pm$0.45      & 55.50$\pm$0.67            & 
		57.41$\pm$0.42            \\
		GcGPN-split                                                  & 
		68.69$\pm$0.43     & 71.83$\pm$0.62 & 57.87$\pm$0.38          & 
		57.78$\pm$0.46      & 57.96$\pm$0.67            & 
		57.36$\pm$0.39            \\
		GcGPN-aux-split                                              & 
		68.30$\pm$0.45     & 73.31$\pm$0.62 & 58.63$\pm$0.40          & 
		57.93$\pm$0.48      & 59.32$\pm$0.68            & 
		58.12$\pm$0.41            \\
		
		GcGPN-cos-aux & 68.03$\pm$0.43 & 71.22$\pm$0.65 & 57.41$\pm$0.41 & 60.26$\pm$0.48 & 54.56$\pm$0.72 & 56.66$\pm$0.45 \\
		
		\bottomrule
	\end{tabular}%
}
\vspace{7pt}
\caption{Test set accuracies (in \%) for $ 5^{+} $-way $ 1 $-shot and $ 5^{+} $-way $ 
	5 $-shot classification on \textit{miniImageNet}.}%
\label{tab:mIN}
\vspace{-1.5em}
\end{table}

\paragraph{Evaluation protocol}
We follow the episodic testing evaluation protocol from previous meta-learning work in (G)FSL~\cite{vinyals2016matching,snell2017prototypical,finn2017model,ravi2016optimization,garcia2017few,gidaris2018dynamic,sung2018learning}
and evaluate all models across 600 test episodes,
where each test episode is an $N^+$-way classification task 
with \emph{all} seen classes plus $N$ novel classes randomly sampled 
from a larger test set containing more than $N$ novel test 
classes.\footnote{Note that in the meta-learning setup, $N$ is usually smaller than 
the number of \emph{all} available novel test classes since the 
label spaces should vary during episodic training.}
The average performance 
with
the 95\% confidence interval 
is
 reported in Table~\ref{tab:mIN} and~\ref{tab:CUB}.
In addition to the evaluation measures Seen-Seen, Novel-Novel and Joint-Joint in~\cite{gidaris2018dynamic}, 
we adopt the convention in GZSL~\cite{xian2017zero} and report
Seen-Joint and Novel-Joint accuracies 
and their harmonic mean,
which capture the \emph{joint} label space classification performance separately for 
seen and novel classes, and the harmonic mean balances the unequal sizes 
of seen and novel classes.
See the supplementary for details on the performance measures and 
pseudo-code for meta-testing.

\paragraph{Baselines}
Recall the
three major requirements for GFSL models:
\textit{(1)} handle dynamic novel label space on-the-fly, \textit{(2)} store 
and represent all seen classes at test time, and \textit{(3)} consistently 
embed 
novel classes into the existing label space.
Most FSL models satisfy (1) but cannot be easily extended to (2). Either the 
entire training set would have to be stored at test time (e.g.~\cite{garcia2017few,vinyals2016matching}), or the model is tailored to $N$-way 
classification only (e.g.~\cite{finn2017model,bauer2017discriminative}). In 
contrast, PN~\cite{snell2017prototypical} 
offers a straight-forward extension PN$^+$ to 
handle requirement (2) as discussed in~\ref{PN+}. Since our paper aims at 
improving GFSL performance, the relevant baselines are PN$^+$ and DFSLwoF~\cite{gidaris2018dynamic}. For the sake of completeness, we compare the 
Novel-Novel accuracy of a GFSL to the performance of FSL models in the 
supplementary.

\paragraph{Model setup for GcGPN}%
\label{sec:model_setup_experiments}
To evaluate the ability 
of leveraging side information for relational 
GFSL, we explore multiple variants of GcGPN with different specifications
for the graph-convolution block.
At the core of almost all model variants is the \emph{semantic} operator $B$ 
containing all graph edge weights (
similarities among all $N_{\text{seen}}+N$ classes).
For reproducibility details on network architecture and hyperparameters 
see the supplementary.
We exploit the model's flexibility to combine multiple operators and include 
variants where the operator set $\mathcal A$ is augmented by the two auxiliary 
operators $\hat B_1$ and $\hat B_2$
defined in eq.~\eqref{eq:auxiliary_ops} (variant indicated by \emph{-aux}).
This allows the model to trade-off between self-connection and the effect of 
similar prototypes.
Further, note that the operators have an inherent four-block structure with 
relations between seen-seen, seen-novel, novel-seen and novel-novel 
classes 
(similar to eq.~\eqref{eq:auxiliary_ops}).
We explore the effect of either utilizing only one semantic operator $\mathcal 
A = \{B\}$ 
with all
class similarities or splitting $B$ into four individual 
operators $\mathcal A = \{B_\text{ss},B_\text{sn},B_\text{ns},B_\text{nn}\}$
with one activated block each. The latter variant, indicated by \emph{-split},
allows the model to learn specialized post-convolution transforms for each 
block.

To further study the effect of the semantic operator and 
the post-convolution transform, we conduct two more ablation experiments on 
\textit{CUB}:
Variant GcGPN-aux-sn has reduced capacity in the operator by deactivating all 
inter-class relations
other than the seen-novel block, whereas GcGPN-aux-fc$\theta_B$ has increased 
capacity in the post-convolution
transform by using fully connected instead of diagonal $\theta_B$.

We also explore GcGPN-cos-aux, 
a very simple way to make use of inter-class relationship modeling 
without using any side information.
We obtain the operator entries by computing cosine similarity between the 
respective class prototypes (see~\ref{sec:GcGPN_model_overview}, graph-conv. operators~(1)).
This also serves as an ablation to understand the effect of the graph-convolution 
based framework without the additional benefit of including side information.
We provide an ablation study on different variants of this in the supplementary, 
including using L2-distance instead of cosine similarity and dropping the auxiliary operators.
For all variants, we use one graph-convolution layer and diagonal post-convolution transform 
$\theta_B$ with learnable entries.

\paragraph{Results and Discussion}
Tables~\ref{tab:mIN} and~\ref{tab:CUB} show results for generalized $5^{+}$-way $K$-shot classification on \textit{miniImageNet} (\textit{mIN}) and \textit{CUB}. 
Since PN$^+$
is only trained for~FSL, its performance 
on novel class queries 
drops drastically when changing 
from the novel to the joint label space (\emph{cf.}~Novel-Novel and 
Novel-Joint).
The novel classes are well-separated from each other 
but not 
consistently embedded into the seen label space. 

GcGPN-cos-aux is the simplest variant with an explicit inter-class relationship model
given by the cosine similarity between class prototypes.
DFSLwoF~\cite{gidaris2018dynamic} also relies on cosine similarity, but between \emph{keys}.
More precisely, every class has a $d$-dimensional key $k_n$, 
which are trainable model parameters \emph{in addition} to the prototypes. 
Thus, DFSLwoF has higher modeling capacity and flexibility for the 
inter-class relations than GcGPN-cos-aux. 
While maintaining an edge on \emph{mIN},
it is clearly outperformed by GcGPN-cos-aux on \emph{CUB} 
in terms of both Joint-Joint accuracy and the harmonic mean.
This shows that our graph-convolution based framework with an in general fully-connected graph
can potentially obtain better performance with a much simpler inter-class relationship model.

\begin{table}[t]
\centering
\setlength{\tabcolsep}{3pt}
\resizebox{0.78\linewidth}{!}{
	\begin{tabular}{ l c c c c c c }
		\toprule[0.25ex]
		\multirow{2}{*}{}                                            
		&           & 
		FSL            &                                   
		\multicolumn{4}{c}{GFSL}                                   \\
		\cmidrule(r){3-3}\cmidrule(r){4-7}\multicolumn{1}{l}{1-shot} & 
		Seen-Seen & 
		Novel-Novel    & Joint-Joint             & Seen-Joint & 
		Novel-Joint               & H-Mean                    \\
		\midrule
		PN$^+$~\cite{snell2017prototypical}                         & 
		35.16$\pm$0.42     & 58.87$\pm$0.91 & 17.61$\pm$0.21          & 
		35.16$\pm$0.42      & \hphantom{0}0.05$\pm$0.02 & 
		\hphantom{0}0.09$\pm$0.04 \\
		DFSLwoF~\cite{gidaris2018dynamic}                            & 
		47.02$\pm$0.56     & 59.87$\pm$0.93 & 37.87$\pm$0.48          & 
		41.55$\pm$0.56      & 34.19$\pm$0.82            & 
		36.34$\pm$0.56            \\
		GcGPN                                                        & 
		43.96$\pm$0.55     & 70.49$\pm$0.81 & 45.46$\pm$0.48          & 
		34.92$\pm$0.54      & 56.00$\pm$0.84            & 
		42.21$\pm$0.47            \\
		GcGPN-aux                                                    & 
		46.26$\pm$0.57     & 71.17$\pm$0.79 & \textbf{47.61$\pm$0.47} & 
		36.35$\pm$0.56      & 58.88$\pm$0.78            & 
		44.21$\pm$0.48            \\
		GcGPN-split                                                  & 
		40.60$\pm$0.53     & 71.77$\pm$0.81 & 46.09$\pm$0.48          & 
		30.49$\pm$0.52      & 61.68$\pm$0.80            & 
		40.15$\pm$0.50            \\
		GcGPN-aux-split                                              & 
		50.99$\pm$0.53     & 71.51$\pm$0.75 & 47.33$\pm$0.46          & 
		45.64$\pm$0.53      & 49.01$\pm$0.77            & 
		46.53$\pm$0.47            \\
		
		GcGPN-cos-aux & 51.79$\pm$0.55 & 59.80$\pm$0.95 & 44.06$\pm$0.52 & 41.25$\pm$0.57 & 46.87$\pm$0.88 & 42.90$\pm$0.52 \\
		
		\midrule
		{Ablations}                                    &           
		&                
		&                         &            &                           
		&                           \\
		\addlinespace[-0.1em]\midrule
		GcGPN-aux-fc$\theta_B$                                       & 
		51.88$\pm$0.55     & 72.72$\pm$0.80 & 47.49$\pm$0.46          & 
		47.33$\pm$0.55      & 47.66$\pm$0.74            & 
		\textbf{46.77$\pm$0.48  
		} \\
		GcGPN-aux-sn                                                 & 
		38.71$\pm$0.56     & 70.25$\pm$0.84 & 44.67$\pm$0.48          & 
		29.26$\pm$0.54      & 60.09$\pm$0.81            & 
		38.61$\pm$0.52            \\
		\midrule[0.25ex]
		\multicolumn{1}{l}{5-shot}                                   & 
		Seen-Seen & 
		Novel-Novel    & Joint-Joint             & Seen-Joint & 
		Novel-Joint               & H-Mean                    \\
		\midrule
		PN$^+$~\cite{snell2017prototypical}                         & 
		43.04$\pm$0.44     & 75.81$\pm$0.67 & 25.26$\pm$0.26          & 
		42.90$\pm$0.44      & \hphantom{0}7.62$\pm$0.32 & 
		12.45$\pm$0.44            \\
		DFSLwoF~\cite{gidaris2018dynamic}                            & 
		48.37$\pm$0.55     & 74.73$\pm$0.79 & 44.97$\pm$0.51          & 
		45.09$\pm$0.55      & 44.85$\pm$0.82            & 
		44.19$\pm$0.54            \\
		GcGPN                                                        & 
		44.33$\pm$0.53     & 76.98$\pm$0.75 & 50.35$\pm$0.46          & 
		36.44$\pm$0.53      & 64.26$\pm$0.75            & 
		45.92$\pm$0.48            \\
		GcGPN-aux                                                    & 
		50.73$\pm$0.56     & 75.87$\pm$0.74 & \textbf{50.62$\pm$0.49} & 
		45.92$\pm$0.54      & 55.33$\pm$0.79            & 
		\textbf{49.53$\pm$0.48  
		} \\
		GcGPN-split                                                  & 
		52.31$\pm$0.53     & 76.49$\pm$0.74 & 49.16$\pm$0.48          & 
		48.37$\pm$0.54      & 49.95$\pm$0.78            & 
		48.42$\pm$0.49            \\
		GcGPN-aux-split                                              & 
		51.39$\pm$0.56     & 76.63$\pm$0.75 & 48.87$\pm$0.50          & 
		47.79$\pm$0.57      & 49.95$\pm$0.80            & 
		48.11$\pm$0.52            \\
		
		GcGPN-cos-aux & 50.56$\pm$0.56 & 74.70$\pm$0.77 & 46.90$\pm$0.48 & 46.82$\pm$0.57 & 46.99$\pm$0.80 & 46.06$\pm$0.50 \\
		
		\midrule
		{Ablations}                                    &           
		&                
		&                         &            &                           
		&                           \\
		\addlinespace[-0.1em]\midrule
		GcGPN-aux-fc$\theta_B$                                       & 
		42.27$\pm$0.54     & 77.38$\pm$0.76 & 50.11$\pm$0.48          & 
		34.21$\pm$0.52      & 66.02$\pm$0.80            & 
		44.45$\pm$0.49            \\
		GcGPN-aux-sn                                                 & 
		45.42$\pm$0.55     & 76.27$\pm$0.74 & 49.37$\pm$0.49          & 
		38.45$\pm$0.55      & 60.29$\pm$0.83            & 
		46.22$\pm$0.48            \\
		\bottomrule
	\end{tabular}%
}
\vspace{7pt}
\caption{Test set accuracies (in \%) for $ 5^{+} $-way $ 1 $-shot and $ 5^{+} $-way $ 
	5 $-shot classification on \textit{CUB}.}%
\label{tab:CUB}
\vspace{-1.5em}
\end{table}

On \textit{mIN}, GcGPN benefits from auxiliary operators and splitting on both tasks. 
Our best variant achieves state-of-the-art 
Joint-Joint accuracy and harmonic mean on the 1-shot task while being competitive with 
DFSLwoF~\cite{gidaris2018dynamic} on the 5-shot task. 
On \textit{CUB}, our model outperforms state-of-the-art by a wide margin 
on both 1-shot and 5-shot tasks and in terms of both Joint-Joint accuracy
and harmonic mean performance. 
These improvements mainly stem from the model's enhanced ability to incorporate 
novel classes 
consistently into the seen class label space, which is suggested by 
the significant increase in Novel-Joint accuracy
while the Seen-Joint accuracy remains comparable with~
\cite{gidaris2018dynamic}.
Comparing to GcGPN-cos-aux, we see that side information has a clear
beneficial effect on the accuracy of around $3\%$.
Unlike on \textit{mIN}, splitting was not beneficial. 
We do observe improvements from using auxiliary operators, however, 
the simplest GcGPN already outperforms the baselines significantly.
Note that our model variants do not require learning an additional key space and
an attention module as in DFSLwoF, but instead relies on side information only.
Thus, the quality of the side information becomes crucial.
The attributes on \textit{CUB} provides fine-grained visual information 
which, according to our empirical results, 
proves to be a richer source of relational information compared to the taxonomy-based 
similarity for \textit{mIN}.

We conducted a further ablation study for GcGPN on \textit{CUB},
which suggests that increasing the post-convolution transformation 
capacity (GcGPN-aux-fc$\theta_B$) improves the model's discriminative power in the 1-shot task. 
Restricting the relational graph to novel-seen dependencies turns out to harm the performance,
which is in line with our key intuition that learning prototypes jointly by
incorporating \emph{all} inter-class relationships helps to handle the challenging trade-off in GFSL.


\section{Conclusion and future work}


We propose GcGPN 
which
takes inter-class relationships defined by weighted graphs 
into account to consistently embed previously seen and novel classes
into a joint prototype space. This allows for better generalization to 
novel tasks while 
maintaining discriminative power 
over not only novel but also all seen classes. Our model generalizes
existing approaches in FSL and GFSL and achieves strong state-of-the-art results
by leveraging side information. 

Future work along this line would include an extensive analysis and
comparison of different kinds of operators. Further, identifying
useful external sources of side information would greatly leverage
the benefits of using semantic operators for few-shot learning tasks.

\newpage
\bibliographystyle{unsrt}
\bibliography{ms}

\appendix
\newpage
\section*{Supplementary Material: Relational Generalized Few-Shot Learning}%

\section{Implementation details}%
\label{supsec:implementation_details}

\subsection{Reproducibility details}
For the sake of reproducibility, we provide comprehensive implementation 
details of our method in this section.
Figure~\ref{fig:gcgpn} depicts the full pipeline of our framework and 
Algorithm~\ref{alg:gcgpn-algo} provides the step-by-step recipe how  
our model GcGPN is used 
to perform GFSL. 

\begin{algorithm}
	\caption{$N^+$-way $K$-shot classification with GcGPN}%
	\label{alg:gcgpn-algo}
	\begin{algorithmic}[1]
		\State{Input: $N_{\text{seen}}, N, \mathcal{A} := \left\{ B_{1}, B_{2}, 
		B_{3}, \ldots \right\}$} \Comment{Number of classes, number of shots, 
		semantic operators}
		\item[]\vspace{-1.9ex}
		\State{Initialize parameters: $\psi, c_1,\ldots,c_{N_{\text{seen}}}, 
		\theta_B, s_B, \tau, \forall B \in \mathcal A$
		}\label{alg:inline-parameters} 
		\item[]\vspace{-1.9ex}
		\For{\( episode = 1, 2, \ldots \)}
		\State{$\mathcal{Y}_{\text{novel}}$, $\mathcal{Y}_{\text{seen}}$, 
		$\mathcal{D}_{\text{novel}}$, $\mathcal{Q}_{\text{joint}}$ $\leftarrow$ 
		Algorithm~\ref{alg:gfsl-episodic-sampling}} \Comment{Sample a $N^+$-way 
		$K$-shot episode}
		\item[]\vspace{-1.9ex}
		\State{$z_{n,i} \leftarrow f_\psi(\mathcal{D}_{\text{novel}}), 
		n=N_{\text{seen}}+1,\ldots,N_{\text{seen}}+N, i=1,\ldots,K$} 
		\Comment{Apply feature extractor $f_\psi$ to all support sets}
		\item[]\vspace{-1.9ex}
		\State{$Z \leftarrow f_\psi(\mathcal{Q}_{\text{joint}})$} \Comment{Apply 
		feature extractor $f_\psi$ to all query examples}
		\item[]\vspace{-1.9ex}
		\State{$c_n \leftarrow \frac{1}{K} \sum_{i=1}^{K}\bar{z}_{n,i}, 
		n=N_{\text{seen}}+1,\ldots,N_{\text{seen}}+N$} \Comment{Average 
		normalized support sets to initial novel prototypes}
		\item[]\vspace{-1.9ex}
		\State{$C \leftarrow 
		(c_1,\ldots,c_{N_{\text{seen}}},c_{N_{\text{seen}}+1},\ldots,c_{N_{\text{seen}}+N})$}
		 \Comment{Concatenate seen and novel initial prototypes}
		\item[]\vspace{-1.9ex}
		\State{$C'=(c'_1,\ldots,c'_{N_{\text{seen}}},c'_{N_{\text{seen}}+1},\ldots,c'_{N_{\text{seen}}+N})
		 \leftarrow \tilde{g}(C, \{\theta_B, \mathcal A, s_B\}_{B \in \mathcal 
		{A}})$} \Comment{Update all prototypes with graph-conv.}
		\item[]\vspace{-1.9ex}
		\State{$p(y=n \mid x, c'_1,\ldots,c'_{N_{\text{seen}}+N}) \leftarrow 
		\frac{\exp\left(\tau \cos(z, 
		c'_n)\right)}{\sum_{m=1}^{N_{\text{seen}}+N} \exp\left(\tau \cos(z, 
		c'_m)\right)}, \forall z \in Z$} \Comment{Predict class probabilities 
		for all queries}
		\item[]\vspace{-1.9ex}
		\If{training}
		\item[]\vspace{-1.9ex}
		\State{Compute loss $L$, take gradient $\delta L$ w.r.t.\ parameters, 
		perform SGD update}
		\item[]\vspace{-1.9ex}
		\State{Adjust learning rate, check early stopping}
		\item[]\vspace{-1.9ex}
		\EndIf{}
		\item[]\vspace{-1.9ex}
		\EndFor{}
	\end{algorithmic}
\end{algorithm}

\textbf{Sampling of GFSL episodes:} The sampling at training time is given in 
Algorithm~\ref{alg:gfsl-episodic-sampling}. At test time, instead of simulating 
novel and seen classes from $\mathcal{Y}_{\text{train}}$, the seen label space is 
given by all training classes, e.g. $\mathcal{Y}_{\text{seen}}=\mathcal{Y}_{\text{train}}$, 
while $\mathcal{Y}_{\text{novel}}$ and $\mathcal{D}_{\text{novel}}$ are sampled from a larger 
test set of novel classes.\\

\textbf{Application of the feature extractor:} The feature extractor $f_\psi$ maps from 
input space into a $d$-dimensional feature space. 
For comparability, we adopt the same feature extractor architecture as in~\cite{snell2017prototypical} 
and~\cite{gidaris2018dynamic} with 4 convolutional blocks and 128 output 
feature maps, 
where each block consists of a $3 \times 3$ convolution layer followed by batch 
normalization, ReLU and $2 \times 2$ max-pooling.\\

\textbf{Initial prototypes:} Seen class initial prototypes $c_n \in 
\mathbb{R}^d, \: n=1,\ldots,N_{\text{seen}}$ are parameters of the model.
Novel class initial prototypes are given by the per-class average 
$c_n = \frac{1}{K} \sum_{i=1}^{K}\bar{z}_{n,i}$ of the normalized support set 
examples 
$\bar{z}_{n,k} = \frac{f_\psi(x_{n,k})}{||f_\psi(x_{n,k})||}$,
$n=N_{\text{seen}}+1,\ldots,N_{\text{seen}}+N$ with $x_{n,k}$ denoting the 
$k$-th labeled support set examples of class $n$.
The $(N_{\text{seen}}+N) \times d$ matrix 
$C=(c_1,\ldots,c_{N_{\text{seen}}},c_{N_{\text{seen}}+1},\ldots,c_{N_{\text{seen}}+N})$
contains all initial prototypes 
with the upper block corresponding to seen and the lower block to novel classes.\\

\textbf{Obtaining operators:} As discussed in 
sec.~\ref{sec:GcGPN_model_overview}, a 
set $\mathcal A$ of operators can be extracted from different kinds of 
inter-class relations.
Here, we describe how the operators used in our experiments are obtained. 
As mentioned in sec.~\ref{sec:model_setup_experiments}, the semantic operator 
$B$ is at the core of all model variants we evaluated.

For \textit{miniImageNet}, the $i,j$-th entry is obtained as \textit{WordNet} 
path similarity between class $i$ and class $j$. 
More precisely, we use the \texttt{path\_similarity} method from the NLTK 
library~\cite{pedersen2004wordnet} with default parameters. 
This measures class similarities based on the shortest path distances between 
the class labels in the \textit{WordNet} taxonomy.
Fig.~\ref{fig:op_sem_min_large} visualizes such a semantic operator on an 
example $5^+$-way episode.

For \textit{CUB}, the $i,j$-th entry is obtained as pairwise cosine similarity 
between class-level attributes,
which are 312-dimensional vectors describing visual characteristics of the 
respective bird species.
These attributes are annotations that come together with the dataset.

We normalize the rows of the semantic operators by a softmax with learnable 
temperature (initialized to 1.0).
Fig.~\ref{fig:op_sem_cub_large} visualizes such a semantic operator on an 
example $5^+$-way episode.
For the semantic-only model variant of GcGPN, $\mathcal{A}=\{B\}$. 
For the \emph{-split} variant, we split $B$ into four individual operator 
$\mathcal A = \{B_\text{ss},B_\text{sn},B_\text{ns},B_\text{nn}\}$
with one activated block each. This is to study the effect of learning 
specialized post-convolution transforms for each block.
For the \emph{-aux} variant, we additionally include auxiliary operators $\hat 
B_1$ and $\hat B_2$ defined in eq.~\eqref{eq:auxiliary_ops}, 
e.g. $\mathcal{A}=\{B, \hat B_1, \hat B_2\}$ or $\mathcal A = \{B_\text{ss}, 
B_\text{sn},B_\text{ns},B_\text{nn}, \hat B_1, \hat B_2\}$.
This is to study the effect of the ability to trade-off between self-connection 
and neighboring prototypes.\\

\textbf{Application of graph convolution:} We apply graph convolution $\tilde{g}$ to the 
initial prototypes $C$ to obtain the updated prototypes
$C'=\tilde{g}(C)=\tilde g^L(\ldots\tilde g^1(C))$, where $g^l,l=1,\ldots,L$ is 
defined in 
sec.~\ref{sec:GcGPN_model_overview}.
In our experiments we use one graph-convolution layer and study two variants 
for the post-convolution transforms $\theta_B$:
either as diagonal matrix or as full matrix (latter variant indicated by 
\emph{-fc$\theta_B$}) with learnable entries.
In both cases, $\theta_B$ is initialized to be the identity matrix.\\

\textbf{Performing classification:} For each query $x\in \mathcal{Q}_{\text{joint}}$, we 
obtain conditional class probabilities 
using the updated prototypes $C'$ according to eq.~\eqref{eq:GcGPN}.
At training time, we compute the cross-entropy loss on these softmax 
probabilities and take the gradient to update all trainable parameters of the 
model.\\

\textbf{Training:} The learnable parameters of GcGPN include the weights $\psi$ 
of the feature extractor,
the seen class initial prototypes $c_1,\ldots,c_{N_{\text{seen}}}$,
the weights of the post-convolutional transform $\theta_B$ and the 
corresponding scaling factor $s_B$ for each operator $B \in \mathcal A$,
temperatures in any operator normalization (if applicable) and the cosine 
classifier.
All parameters are learned end-to-end and the model trained from scratch,
unlike the two-phase training used in~\cite{gidaris2018dynamic} or approaches 
using pre-trained image features such 
as~\cite{verma2017simple,schonfeld2018generalized,xian2019f}.
All models are trained for 75 epochs on \textit{miniImageNet} and 45 epochs on 
\textit{CUB} 
using an SGD optimizer with a momentum of 0.9, 
a weight decay parameter of 0.0005 and an initial learning rate of 0.1 that was 
reduced after 20, 40, 50 and 60 epochs. 
Performance is monitored on the validation set for early stopping.

\subsection{Pseudo code for GFSL episodic sampling}%
\label{supsec:pseudo_code}

\newcommand{\rs}{\textsc{RandomSample}}
We provide pseudo code for episodic sampling at training time under the GFSL 
setup in Algorithm~\ref{alg:gfsl-episodic-sampling}.
For test time evaluation, instead of simulating novel and seen classes from 
$\mathcal{Y}_{\text{train}}$, 
the seen label space is given by all training classes, e.g. 
$\mathcal{Y}_{\text{seen}}=\mathcal{Y}_{\text{train}}$, 
while $\mathcal{Y}_{\text{novel}}$ and $\mathcal{D}_{\text{novel}}$ are sampled from a larger 
test set of novel classes.
\begin{algorithm*}[t]
	\caption[GFSL episodic sampling]{Sampling of an $N^+$-way $K$-shot episode 
	from a set of training data \( \mathcal{D}_{\text{train}} \), where \( 
	\mathcal{D}_{}^{(i)} \) only contains elements of class \( i \). \( N \) 
	denotes the number of novel classes per episode, \( K \) the number of 
	instances in the support set, \( Q \) the number of query instances and 
	finally \( B \) the number of instances per seen class.
		\( \rs(S,N) \) describes uniform random sampling of \( N \) elements 
		without replacement from a set \( S \)}%
	\label{alg:gfsl-episodic-sampling}
	\begin{algorithmic}[1]
		\State{Input: $N_{\text{train}}$, $N$, $K$, $Q$, $B$}
		\item[]\vspace{-1.9ex}
		\State{\( \mathcal{Y}_{\text{novel}} \leftarrow \textsc{\rs}(\{1, \ldots, 
		N_{\text{train}}\}, N)\)} \Comment{Sample \enquote{fake} classes for episode}
		\item[]\vspace{-1.9ex}
		\State{ \( \mathcal{Y}_{\text{seen}} \leftarrow \{1, \ldots, N_{\text{train}}\} 
		\setminus \mathcal{Y}_{\text{novel}} \) } \Comment{Store remaining seen 
		classes}
		\item[]\vspace{-1.9ex}
		\For{\( i \) in \( \mathcal{Y}_{\text{novel}} \)}
		\item[]\vspace{-1.9ex}
		\State{\( D_{\text{novel}}^{(i)} \leftarrow 
		\rs\left(\mathcal{D}_{\text{train}}^{(i)}, K\right) \)} \Comment{Sample 
		support instances}
		\item[]\vspace{-1.9ex}
		\State{\( \mathcal{Q}_{\text{novel}}^{(i)} \leftarrow 
		\rs\left(\mathcal{D}_{\text{train}}^{(i)} \setminus D_{\text{novel}}^{(i)}, Q\right) 
		\)} \Comment{Sample novel query instances}
		\item[]\vspace{-1.9ex}
		\EndFor{}
		\item[]\vspace{-1.9ex}
		\For{\( j \) in \( \mathcal{Y}_{\text{seen}} \)}
		\item[]\vspace{-1.9ex}
		\State{\( \mathcal{Q}_{\text{seen}}^{(j)} \leftarrow 
		\rs(\mathcal{D}_{\text{train}}^{(j)}, B) \)} \Comment{Sample seen query 
		instances}
		\item[]\vspace{-1.9ex}
		\EndFor{}
		\item[]\vspace{-1.9ex}
		\State{\( \mathcal{Q}_{\text{joint}} \leftarrow \mathcal{Q}_{\text{novel}} \cup 
		\mathcal{Q}_{\text{seen}} \)}
		\item[]\vspace{-1.9ex}
		\State{Output: $\mathcal{D}_{\text{novel}}$, $\mathcal{Q}_{\text{joint}}$} 
		\Comment{Output support sets and query sets}
	\end{algorithmic}
\end{algorithm*}

\section{Experimental details}

\subsection{Ablation study for GcGPN without side information}
\label{supsec:GcGPN_opdist}

\begin{table}
	\centering
	\setlength{\tabcolsep}{3pt}
	\resizebox{0.9\linewidth}{!}{
		\begin{tabular}{ l c c c c c c }
			\toprule[0.25ex]
			\multirow{2}{*}{}  	&   & FSL &    \multicolumn{4}{c}{GFSL}\\
			\cmidrule(r){3-3}\cmidrule(r){4-7}\multicolumn{1}{l}{1-shot} & Seen-Seen & Novel-Novel & Joint-Joint & Seen-Joint & Novel-Joint & H-Mean\\
			\midrule
			GcGPN-cos & 65.14$\pm$0.44\% & 55.08$\pm$0.75\% & 47.25$\pm$0.41\% & 54.65$\pm$0.46\% & 39.86$\pm$0.75\% & 45.24$\pm$0.52\% \\
			GcGPN-l2 & 53.22$\pm$0.46\% & 52.45$\pm$0.77\% & 40.14$\pm$0.41\% & 35.70$\pm$0.47\% & 44.59$\pm$0.75\% & 38.83$\pm$0.40\% \\
			GcGPN-cos-aux & 69.86$\pm$0.41\% & 54.00$\pm$0.77\% & 47.94$\pm$0.40\% & 62.39$\pm$0.45\% & 33.50$\pm$0.67\% & 42.88$\pm$0.59\% \\
			GcGPN-l2-aux & 70.08$\pm$0.41\% & 54.46$\pm$0.75\% & 48.21$\pm$0.40\% & 62.78$\pm$0.43\% & 33.65$\pm$0.68\% & 43.09$\pm$0.59\% \\
			\midrule[0.25ex]
			\multicolumn{1}{l}{5-shot} & Seen-Seen & Novel-Novel    & Joint-Joint& Seen-Joint & Novel-Joint   & H-Mean\\
			\midrule
			GcGPN-cos & 55.44$\pm$0.46\% & 68.52$\pm$0.65\% & 50.62$\pm$0.41\% & 43.56$\pm$0.49\% & 57.68$\pm$0.72\% & 49.04$\pm$0.40\% \\
			GcGPN-l2 & 57.53$\pm$0.44\% & 69.25$\pm$0.66\% & 51.17$\pm$0.41\% & 47.98$\pm$0.46\% & 54.36$\pm$0.75\% & 50.29$\pm$0.40\% \\
			GcGPN-cos-aux & 68.03$\pm$0.43\% & 71.22$\pm$0.65\% & 57.41$\pm$0.41\% & 60.26$\pm$0.48\% & 54.56$\pm$0.72\% & 56.66$\pm$0.45\% \\
			GcGPN-l2-aux & 67.79$\pm$0.43\% & 72.37$\pm$0.62\% & 57.81$\pm$0.41\% & 59.30$\pm$0.46\% & 56.32$\pm$0.68\% & 57.29$\pm$0.43\% \\
		\bottomrule
		\end{tabular}
	}
	\vspace{7pt}
	\caption{Test set accuracies for $ 5^{+} $-way $ 1 $-shot and $ 5^{+} $-way $ 
	5 $-shot classification on \textit{mIN}.}%
	\label{tab:opdist_mIN}
\end{table}

\begin{table}
	\centering
	\setlength{\tabcolsep}{3pt}
	\resizebox{0.9\linewidth}{!}{
		\begin{tabular}{ l c c c c c c }
			\toprule[0.25ex]
			\multirow{2}{*}{} &     & 	FSL    &  \multicolumn{4}{c}{GFSL}\\
			\cmidrule(r){3-3}\cmidrule(r){4-7}\multicolumn{1}{l}{1-shot} & Seen-Seen & Novel-Novel & Joint-Joint & Seen-Joint & Novel-Joint & H-Mean\\
			\midrule
			GcGPN-cos & 44.19$\pm$0.56\% & 60.86$\pm$0.93\% & 38.06$\pm$0.48\% & 39.15$\pm$0.56\% & 36.97$\pm$0.84\% & 36.85$\pm$0.51\% \\
			GcGPN-l2 & 44.10$\pm$0.55\% & 60.00$\pm$0.91\% & 38.84$\pm$0.49\% & 36.82$\pm$0.56\% & 40.86$\pm$0.88\% & 37.58$\pm$0.50\% \\
			GcGPN-cos-aux & 51.79$\pm$0.55\% & 59.80$\pm$0.95\% & 44.06$\pm$0.52\% & 41.25$\pm$0.57\% & 46.87$\pm$0.88\% & 42.90$\pm$0.52\% \\
			GcGPN-l2-aux & 45.99$\pm$0.56\% & 60.30$\pm$0.93\% & 41.88$\pm$0.52\% & 35.47$\pm$0.56\% & 48.28$\pm$0.87\% & 40.00$\pm$0.50\% \\
			\midrule[0.25ex]
			\multicolumn{1}{l}{5-shot} & Seen-Seen & Novel-Novel & Joint-Joint & Seen-Joint & Novel-Joint & H-Mean\\
			\midrule
			GcGPN-cos & 43.10$\pm$0.53\% & 74.82$\pm$0.81\% & 45.79$\pm$0.49\% & 37.22$\pm$0.52\% & 54.36$\pm$0.86\% & 43.40$\pm$0.47\% \\
			GcGPN-l2 & 43.05$\pm$0.57\% & 74.47$\pm$0.80\% & 45.82$\pm$0.51\% & 37.23$\pm$0.56\% & 54.41$\pm$0.85\% & 43.42$\pm$0.49\% \\
			GcGPN-cos-aux & 50.56$\pm$0.56\% & 74.70$\pm$0.77\% & 46.90$\pm$0.48\% & 46.82$\pm$0.57\% & 46.99$\pm$0.80\% & 46.06$\pm$0.50\% \\
			GcGPN-l2-aux & 51.47$\pm$0.55\% & 74.65$\pm$0.75\% & 48.20$\pm$0.49\% & 47.52$\pm$0.56\% & 48.88$\pm$0.79\% & 47.44$\pm$0.50\% \\
		\bottomrule
		\end{tabular}}
	\vspace{7pt}
	\caption{Test set accuracies for $ 5^{+} $-way $ 1 $-shot and $ 5^{+} $-way $ 
	5 $-shot classification on \textit{CUB}.}%
	\label{tab:opdist_CUB}
\end{table}

In Section~\ref{sec:GcGPN_model_overview}, we discussed that our framework accommodates
different kinds of operators to model inter-class relationships.
As we mentioned, a simple choice can be any distance or similarity measure
on the prototype space, i.e., the operator entry $B_{m,n}$ is given by $dist(c_m, c_n)$.
Here, we provide experimental results for GcGPN using such simple distance operators,
in particular L2 distance (\emph{GcGPN-l2}) and cosine similarity (\emph{GcGPN-cos}).
Similar to \ref{sec:experiments}, we also consider variants with and without
the auxiliary operators $\hat B_1$ and $\hat B_2$ defined in eq.~\eqref{eq:auxiliary_ops}
(variant indicated by \emph{-aux}).

Table~\ref{tab:opdist_mIN} and~\ref{tab:opdist_CUB} show the results on the
\emph{miniImageNet} and \emph{CUB} datasets, respectively.
Generally, L2 distance seems to have slight advantage over cosine similarity
and the use of auxiliary operators increases performance overall.
As discussed in \ref{PN+}, the competitor model DFSLwoF~\cite{gidaris2018dynamic} 
can be seen as a special case of our framework with 
$\mathcal{A} = \left\{\hat{B}_1, \hat B_2, \hat B_{key} \right\}$
where $\hat B_{key}$ is an operator based on a learned key space
(see \ref{sec:GcGPN_model_overview}, graph-conv. operators (3)).
More precisely, the pairwise class similarities of the GcGPN variants here are computed on
the class prototypes $c_n,\: n=1,...,N_{seen}+N_{novel}$, whereas those of DFSLwoF are
computed between the class keys $k_n,\: n=1,...,N_{seen}+N_{novel}$, which are optimized
model parameters \emph{in addition} to the prototypes. 
Thus, DFSLwoF has higher modeling capacity and flexibility for the inter-class 
relations than our simple GcGPN variants.
While it maintains an edge over GcGPN-*-aux of around 2\% on \emph{miniImageNet},
the GcGPN-*-aux variants outperform it on \emph{CUB} in terms of both Joint-Joint and H-Mean
accuracy with a margin of about 2 to 3\% on the 5-shot task and about 4 to 6\% on the 1-shot task.
This shows that with our framework, we can potentially obtain the same performance 
with a much simpler inter-class relationship model.

\subsection{Comparison to FSL methods}%
\label{supsec:comparisonFSL}

In sec.~\ref{sec:experiments} of the paper, we discussed the major requirements 
for GFSL models, which are 
(1) handle dynamic novel label space on-the-fly,
(2) store and represent all seen classes at test time and
(3) consistently embed novel classes into the existing label space.
Although FSL models can address (1), they cannot be easily extended to cover 
requirements (2) and (3). 
Therefore, sec.~\ref{sec:experiments} focuses on comparing our approach GcGPN 
with relevant \emph{GFSL} methods $PN^+$ (naive extension of 
PN~\cite{snell2017prototypical} to GFSL) and DFSLwoF~\cite{gidaris2018dynamic}. 
Nevertheless, we can compare the \emph{FSL} performance of GFSL models, 
which is captured by the performance measure Novel-Novel, to recent FSL models.
Note that all FSL models are trained with the few-shot objective in 
eq.~\eqref{eq:fsl}, 
whereas the GFSL models (DFSLwoF and GcGPN) are trained with the objective in 
eq.~\eqref{eq:gfsl}.
Table~\ref{tab:FSL_mIN} show the results for $5$-way $1$-shot and $5$-way 
$5$-shot classification on \textit{miniImageNet}.
The numbers suggest that (a) GFSL methods outperform FSL methods even on FSL 
tasks, 
and (b) additionally exploiting inter-class relations further improves 
performance.

\begin{table}
	\centering
	\setlength{\tabcolsep}{3pt}
	\resizebox{0.5\linewidth}{!}{
		\begin{tabular}{ l c c }
			\toprule
			\multirow{2}{*}{}                                     	& 5-way 
			1-shot       & 5-way 5-shot            \\
			\midrule
			Matching Network~\cite{vinyals2016matching}            & 
			46.6\%    			 & 60.0\% \\
			PN~\cite{snell2017prototypical}                        & 
			46.61$\pm$0.78\%     & 65.77$\pm$0.70\% \\
			MAML~\cite{finn2017model}                         		& 
			48.07$\pm$1.75\%     & 63.15$\pm$0.91\% \\
			Meta-LSTM~\cite{ravi2016optimization}                  & 
			43.44$\pm$0.77\%     & 60.60$\pm$0.71\% \\
			Meta-SGD~\cite{li2017meta}                          	& 
			50.47$\pm$1.87\%     & 64.03$\pm$0.94\% \\
			REPTILE~\cite{nichol2018reptile}                        & 
			49.97$\pm$0.32\%     & 65.99$\pm$0.58\% \\
			VERSA~\cite{gordon2018meta}                         	& 
			53.40$\pm$1.82\%     & 67.37$\pm$0.86\% \\
			CAVIA~\cite{zintgraf2019fast}                        	& 
			51.82$\pm$0.65\%     & 65.85$\pm$0.50\% \\
			Relation Net~\cite{sung2018learning}                    & 
			50.44$\pm$0.82\%     & 65.32$\pm$0.70\% \\
			Parameter prediction~\cite{qiao2018few}                    & 
			54.53$\pm$0.40\%     & 67.87$\pm$0.20\% \\
			\midrule
			PN$^+$ (sec.~\ref{sec:experiments})	                   	& 
			53.88$\pm$0.78\%     & 70.84$\pm$0.66\% \\
			DFSLwoF~\cite{gidaris2018dynamic}                    	& 
			55.80$\pm$0.78\%     & 72.59$\pm$0.62\% \\
			GcGPN-cos-aux (ours)                                           	& 
			54.00$\pm$0.77\%     & 71.22$\pm$0.65\% \\
			GcGPN (ours)                                           	& 
			55.67$\pm$0.73\%     & 71.53$\pm$0.63\% \\
			GcGPN-aux (ours)                                       	& 
			56.59$\pm$0.75\%     & 71.81$\pm$0.64\% \\
			GcGPN-split (ours)                                     	& 
			55.68$\pm$0.76\%     & 71.83$\pm$0.62\% \\
			GcGPN-aux-split (ours)                                 	& 
			\textbf{60.40$\pm$0.71\%}     & \textbf{73.31$\pm$0.62\%} \\
			\bottomrule
		\end{tabular}%
	}
\vspace{2ex}
	\caption{FSL performance (Novel-Novel) on $5$-way $1$-shot and $5$-way 
	$5$-shot classification on \textit{miniImageNet}.
		The upper part of the table contains FSL methods and the lower part 
		GFSL methods. 
		Numbers for the FSL models are as reported in~\cite{zintgraf2019fast} 
		and~\cite{sung2018learning}
		and numbers for the GFSL models are obtained from our own experiments.
	}%
	\label{tab:FSL_mIN}
	\vspace*{-2.5ex}
\end{table}

\subsection{Definition of performance measures}%
\label{supsec:def_perf_measures}

For the experimental results in Table~\ref{tab:mIN} and~\ref{tab:CUB} from 
sec.~\ref{sec:experiments},
we report several different performance measures according to the conventions 
in GFSL and GZSL. 
We provide the definitions here:
\textbf{Novel-Novel} measures the accuracy when classifying novel class queries 
in the novel class label space.
\textbf{Seen-Seen} measures the accuracy when classifying seen class queries in 
the seen class label space.
\textbf{Joint-Joint} measures the accuracy when classifying seen and novel 
queries in the joint label space.
\textbf{Novel-Joint} measures the accuracy when classifying novel class queries 
in the joint label space.
\textbf{Seen-Joint} measures the accuracy when classifying seen class queries 
in the joint label space.
\textbf{Harmonic Mean (H-Mean)} is the harmonic mean of Novel-Joint and 
Seen-Joint, 
where $H(x_1,x_2)=\frac{2 \cdot x_1 \cdot x_2}{ x_1 + x_2}$ denotes the 
harmonic mean of two numbers $x_1$ and $x_2$.

The performance measures Novel-Novel, Seen-Seen and Joint-Joint accuracies are 
reported in~\cite{gidaris2018dynamic}.
In addition to them, we also adopt the convention in GZSL (generalized 
zero-shot learning)~\cite{xian2017zero} and report 
Novel-Joint and Seen-Joint accuracies together with their harmonic mean.
As~\cite{xian2017zero} points out, the Novel-Joint performance is of particular 
interest 
because GZSL models often fail here drastically in spite of good Novel-Novel 
performance. 
Further, the harmonic mean is often preferred over Joint-Joint accuracy which 
is easily dominated 
by the seen class performance.
This is because queries are much more likely to stem from the seen classes, thus the Joint-Joint accuracy correlates heavily with the Seen-Seen accuracy.

\subsection{Varying the number of shots}

\begin{figure}[h]
	\centering
	\includegraphics[width=0.75\linewidth]{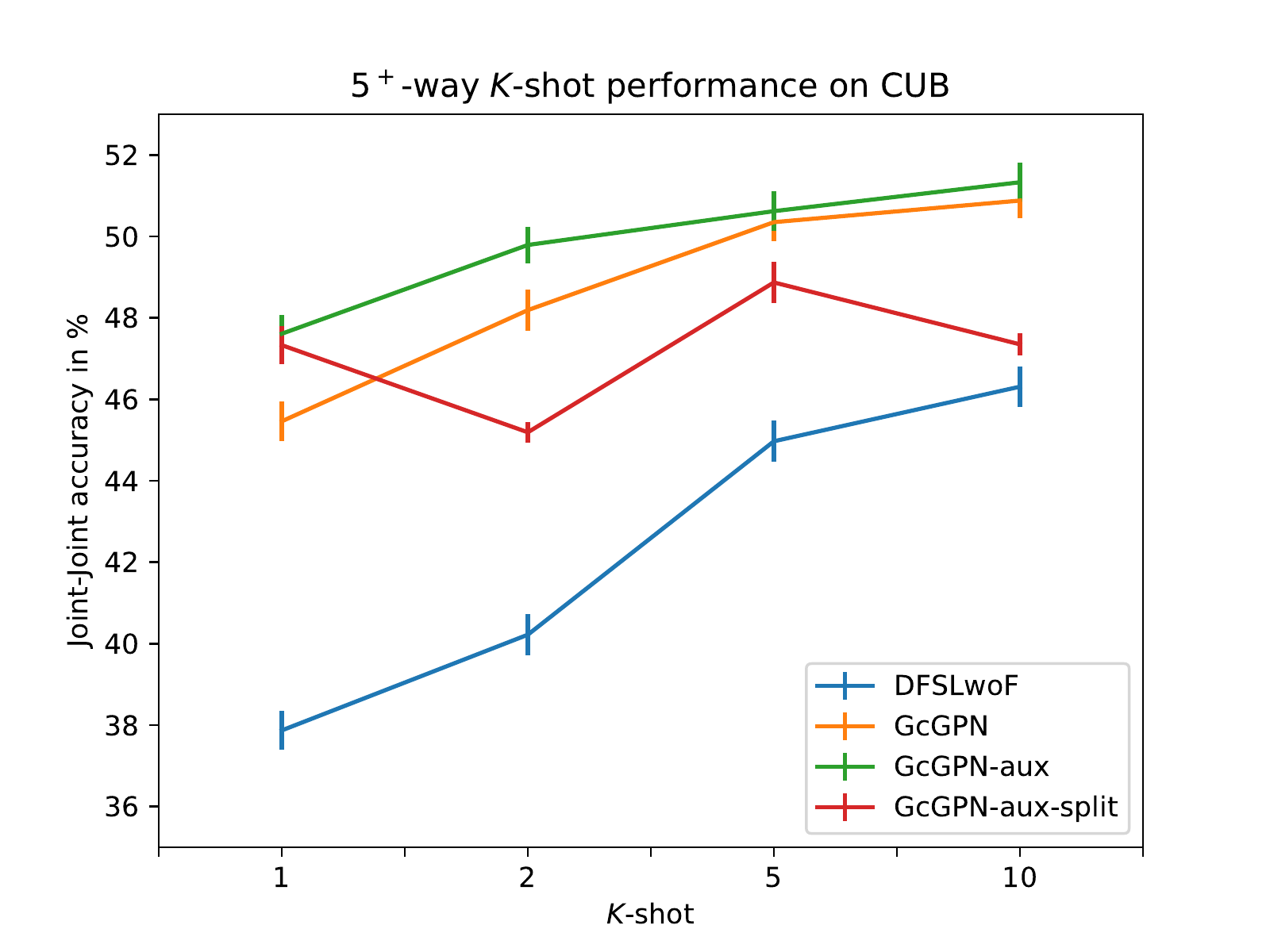}
	\caption{
		$5^+$-way $K$-shot classification accuracy (Joint-Joint) on the 
		\textit{CUB} dataset for different $K$.}%
	\label{fig:kshot}
	\vspace*{-3.5ex}
\end{figure}

We study the model's behavior under different few-shot settings with varying 
numbers $K$ of available labeled examples per novel class.
Fig.~\ref{fig:kshot} shows the Joint-Joint accuracy for $5^{+}$-way $K$-shot 
classification on the \textit{CUB} dataset.
The GcGPN variants are explained in sec.~\ref{sec:model_setup_experiments} 
and~\ref{supsec:implementation_details}.
We train separate models for each $K$ and evaluate their performance.
The results show that the GcGPN variants consistently outperform the baseline 
DFSLwoF~\cite{gidaris2018dynamic}.

\subsection{Semantic operators}
\begin{figure*}[h]
	\centering
	\includegraphics[width=\linewidth,trim=0 0 0 
	59,clip]{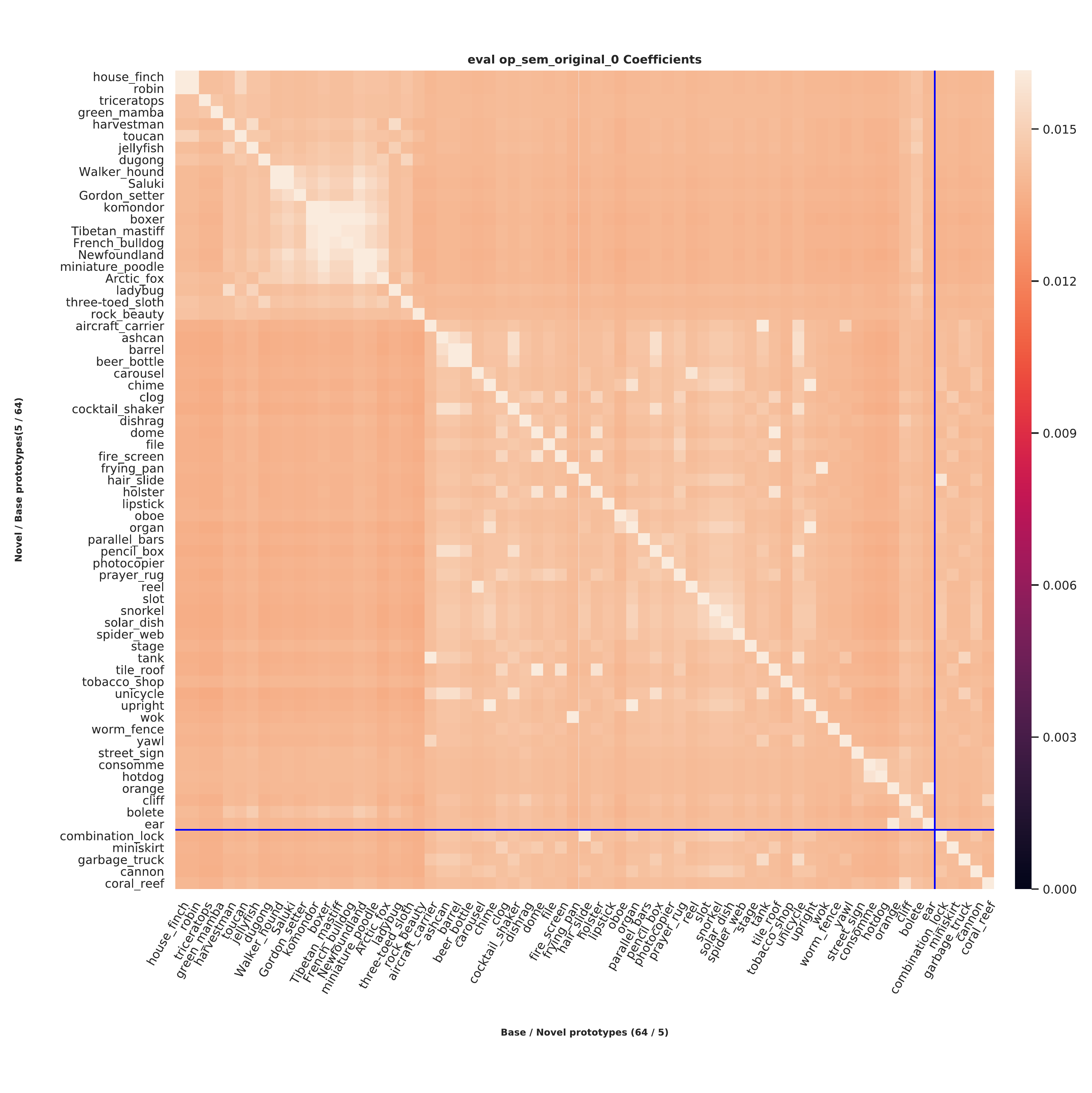}
	\caption{Softmax normalized semantic operators for the 
	\textit{miniImageNet}~\cite{vinyals2016matching} dataset for a typical GFSL 
	$5^{+}$-way $K$-shot task.
		Two large blocks are visible, which indicate the similarities of 
		animate (classes \texttt{house\_finch} to \texttt{three-toed\_sloth}) 
		and inanimate things (remaining classes).
		(Best viewed in color)}%
	\label{fig:op_sem_min_large}
\end{figure*}

\begin{figure*}[h]
	\centering
	\includegraphics[width=\linewidth,trim=0 0 0 
	45,clip]{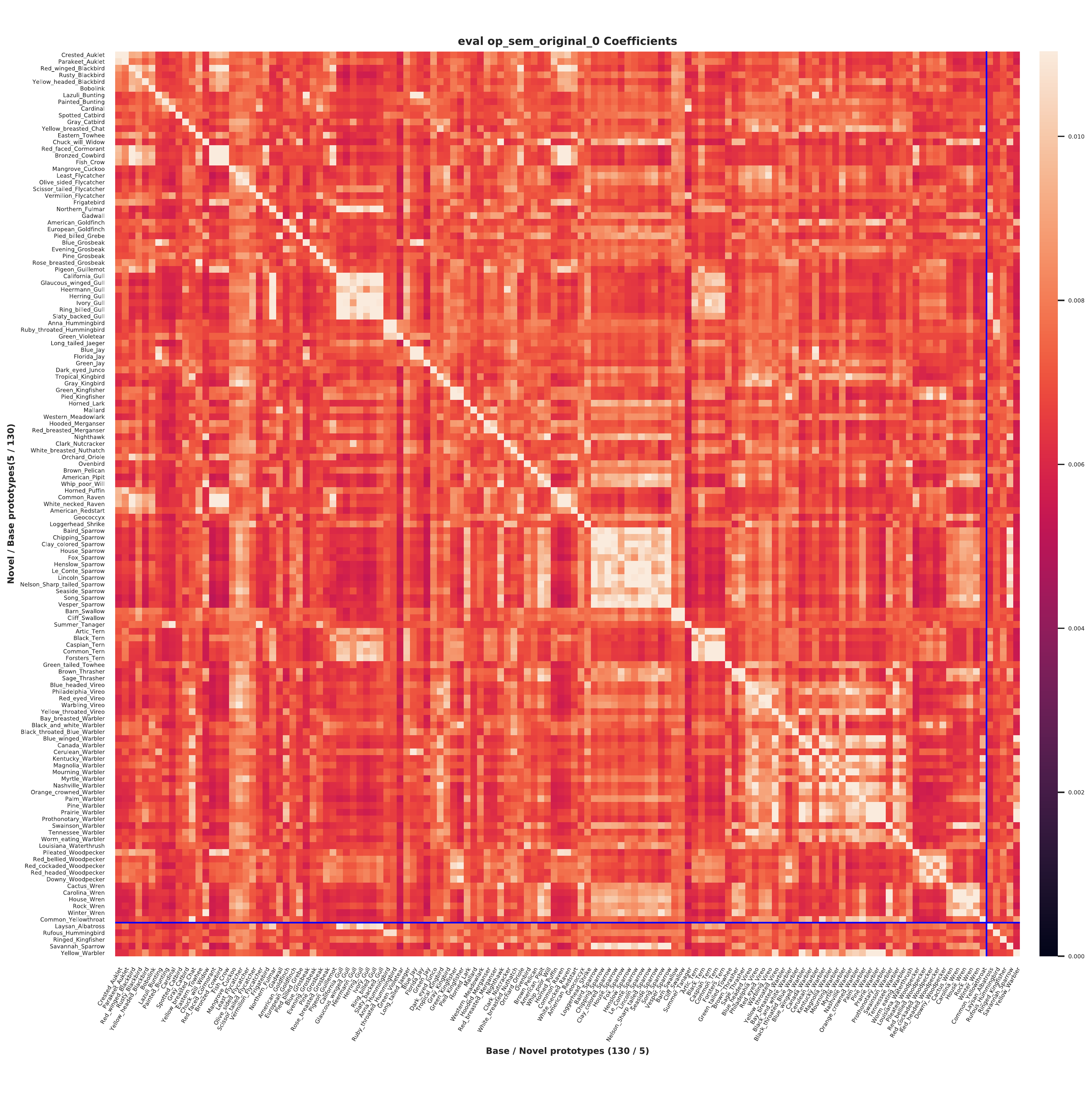}
	\caption{Softmax normalized semantic operators for the Caltech-UCSD 
	Birds-200-2011~(CUB)~\cite{WahCUB_200_2011} dataset for a typical GFSL 
	$5^{+}$-way $K$-shot task.
		The largest continuous block are several different sparrow species 
		(classes \texttt{Baird\_Sparrow} to \texttt{Vesper\_Sparrow}).
		(Best viewed in color)}%
	\label{fig:op_sem_cub_large}
\end{figure*}

In this section we show the semantic operators $B$ for the 
\textit{miniImageNet}~\cite{vinyals2016matching} dataset 
(Figure~\ref{fig:op_sem_min_large}) and for the Caltech-UCSD 
Birds-200-2011~(\textit{CUB})~\cite{WahCUB_200_2011} dataset 
(Figure~\ref{fig:op_sem_cub_large}).
For both visualizations operator temperature is at $1$ leaving the operators 
unmodified. 

Each row and column represents one class. 
In both visualizations, brighter color indicates higher inter-class similarity 
and block 
structures arise when similar classes are listed next to each other. Note that 
the colormap clips the largest values on the diagonal to visualize 
the off-diagonal structure of the side information. Since graph-convolution 
operators usually require normalization, we apply 
row-wise softmax with a learnable temperature such that the entries of each row sum up to 1.

The blue lines divide the operator into four blocks corresponding to 
relations among seen classes in the upper left (Seen-Seen), novel classes in 
the lower 
right (Novel-Novel) and 
mixed relations in the other two blocks (Seen-Novel and Novel-Seen). 

The figures indicate that relational structures are more prominent in 
\textit{CUB} 
as compared 
to \textit{miniImageNet}. 
This reflects that in \textit{WordNet}, the 100 \textit{miniImageNet} classes 
are a small subset from a much larger taxonomy and are almost equally related 
to each other.
In contrast to that, the dedicated fine-grained attributes in \textit{CUB} 
provides
more structural and discriminative information, 
which proves to be particularly beneficial.


%
%
%

\end{document}